\pdfoutput=1
\documentclass[letterpaper, 10 pt, conference]{ieeeconf} 
\IEEEoverridecommandlockouts
\usepackage{times}
\usepackage{amsmath}
\usepackage{amssymb} 
\usepackage{booktabs}
\usepackage{physics}

\usepackage{graphicx}
\usepackage{url}
\usepackage{float}
\usepackage{dblfloatfix}
\usepackage{textcomp}

\usepackage{gensymb}
\usepackage{multicol,tabularx,capt-of}
\usepackage{multirow}

\usepackage{booktabs}
\usepackage{multirow}
\usepackage{svg}
\usepackage{flafter}
\usepackage{multirow}
\usepackage{footnote}
\let\labelindent\relax

\usepackage[shortlabels]{enumitem}

\usepackage{color}

\title{\LARGE \bf
A Systematic Comparison of Simulation Software for Robotic Arm Manipulation using ROS2
}

\author{Florent P. Audonnet$^{1}$, Andrew Hamilton$^{2}$ and Gerardo Aragon-Camarasa$^{1,2}$
\thanks{This research has been supported by EPSRC DTA No. 2605103 and
NVIDIA Corporation for the donation of the Titan Xp GPU.}
\thanks{$^{1}$ School of Computing Science, University of Glasgow, G12 8QQ, Scotland, United Kingdom {\tt\small f.audonnet.1@research.gla.ac.uk}  and {\tt\small gerardo.aragoncamarasa@glasgow.ac.uk}}%
\thanks{$^{2}$ National Manufacturing Institute in Scotland, Scotland, United Kingdom {\tt\small andrew.w.hamilton@strath.ac.uk}}%
}

\begin{document}

\maketitle
\thispagestyle{empty}
\pagestyle{empty}

\begin{abstract}

Simulation software is a powerful tool for robotics research, allowing the virtual representation of the real world. However with the rise of the Robot Operating System (ROS), there are new simulation software packages that have not been compared within the literature. This paper proposes a systematic review of simulation software that are compatible with ROS version 2. The focus is research in robotics arm manipulation as it represents the most often used robotic application in industry and their future applicability to digital twins. For this, we thus benchmark simulation software under similar parameters, tasks and scenarios, and evaluate them in terms of their capability for long-term operations, success at completing a task, repeatability and resource usage. We find that there is no best simulation software overall, but two simulation packages (Ignition and Webots) have higher stability than other while, in terms of resources usage, PyBullet and Coppeliasim consume less than their competitors.
\end{abstract}

\section{INTRODUCTION}

With the advent of deep learning technologies, current research efforts have been focused on teaching robots how to perform various tasks autonomously. However, a data-driven approach is required to acquire and process the vast amount of data to effectively teach a robot how to perform a task which is unfeasible using a real robotic testbed. For this, robot simulation software \cite{ignition,webots,isaac,pybullet,vrep} have been used to overcome the shortcomings of data-hungry AI approaches and to allow the developer to obtain a constant environment~\cite{banks_1999_nodate}. In a simulated environment the world can be controlled, including aspects that would be impractical in reality. There is also no risk of damaging the robot or human operators, and simulations allow to control the time which increases the speed of data collection. 

\begin{figure}[t]
    \centering
    \includegraphics[width=0.45\textwidth]{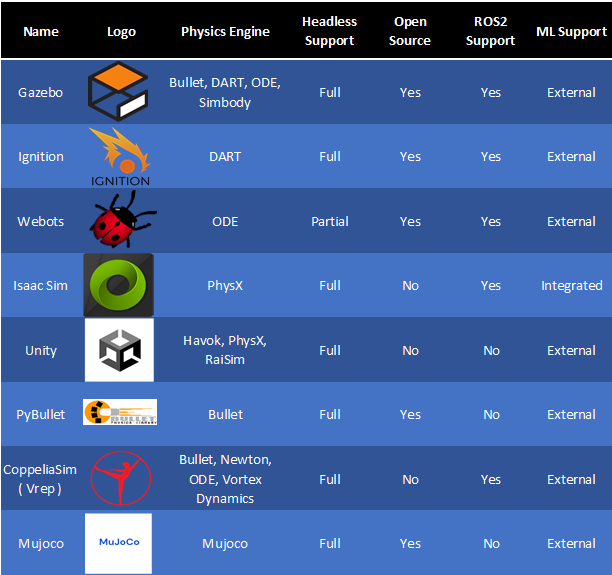}
    \caption{\label{fig:simtable} Overview of the Simulation Software and Their Capabilities }
\end{figure}

Simulations are the gateway for Digital Twins, a high-fidelity representation of the physical world \cite{lu_digital_2020}, and can allow manufacturing to increase production and flexibility of supply chains. Therefore, digital twinning consists of interconnecting a simulation software to a real autonomous robotic system in order to reduce the implementation time of manufacturing process when changing a production line. A recent example of a digital twin solution for a robotic arm can be found in  \cite{tavares_flexible_2018} where the authors used ROS (Robot Operating System)~\cite{ros} to achieve seamless operation between the real and digital world. However, simulation software are not perfect because their physics engines are not an accurate representation of the real world. Furthermore, simulations allow for perfect data capture with no noise which has powered research in deep learning approaches for robotics.

In this paper, we propose to carry out a systematic benchmark of current simulation software (Figure \ref{fig:simtable}) to investigate their performance and suitability to perform different robotic manipulation tasks using the ROS2 (Robot Operating System version 2). ROS has become the de facto communication platform for modern robotic systems. We choose ROS2 because it supports a wide array of devices (e.g. micro-controllers) which enables the integration of Internet of Things (IoT). The latter is a main requirement for developing a working digital twin system. ROS2 can also be used to bridge the gap between AI-enabled robots and real world robot control. We choose robotic arms in this paper as they are prevalent in automated manufacturing operations.

\begin{figure*}[t]
\centering
\includegraphics[width=0.85\textwidth]{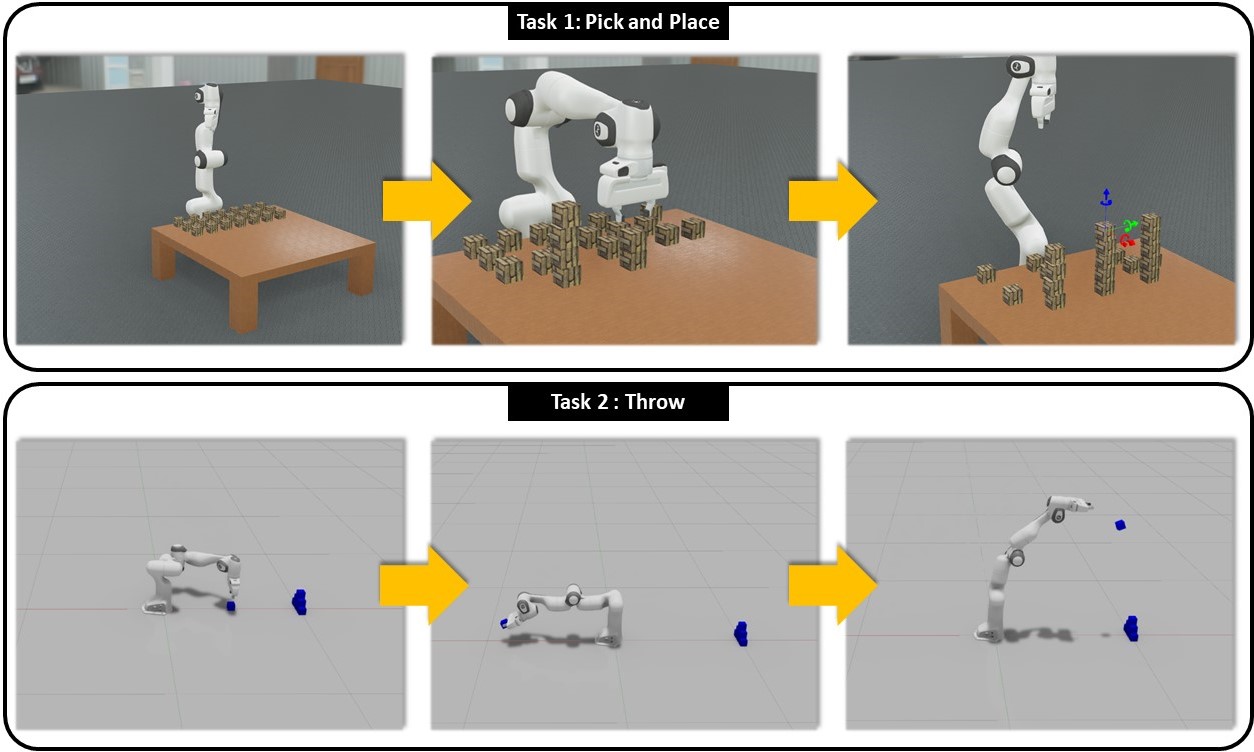}
\caption{\label{fig:overview} Simulation Tasks Progression over Time. Task 1 (top), is a Pick and Place task where the goal is to stack 3 columns of 5 cubes. Task 2 (bottom) is a Throw task where the goal is to collapse a pyramid of 6 cubes by throwing a cube at it. }
\end{figure*}

We consider 2 tasks for the robot arm to perform. The first task is about picking and placing an object which is a common operation in industry. The second task consists of throwing a cube into a pyramid. We chose this throwing task as we aim to test the accuracy and repeatability of the simulation software to decide its potential suitability for building digital twins. Figure \ref{fig:overview} shows an overview of the tasks. We record the resources usage of each simulation considered in this paper while performing each task in both a headless and a graphical version. Our contributions consist of proposing a systematic comparison of state-of-the-art robotic arm simulation software using ROS2, the state-of-the-art version of the robot operating system. Furthermore, we develop an experimental methodology to evaluate robot simulations software for long-term operations and their success at completing a task. We also devised an experimental validation system to evaluate the stability of robot simulation software and their capacity to repeat a given task.


\section{BACKGROUND} \label{sec:background}
Benchmarking robotic simulation software can trace back its origins to Oreback and Christensen \cite{oreback_evaluation_2003}. They were the first to propose a methodology to test robotic simulations. Their approach consisted of summarising the capabilities of 3 simulators, considering quantitative values such as supported OS or programming languages and qualitative opinions such as the learning curve or the difficulty of installation. They also recorded the amount of RAM used while using the simulation software to control a real robot. Kramer and Scheutz \cite{kramer_development_2007} extended \cite{oreback_evaluation_2003} and developed a comprehensive testing suite for open-source robot simulation software. They devised a set of criteria based on the software-development process and created a feature score based on different properties such as usability, supported features (path planning, voice recognition, etc.) and faults handling. They performed a low-level implementation of a task on a real robot and recorded the resource usage. However, the task is scarcely described and was only repeated three times. 

Before ROS \cite{ros}, roboticists used the simulation software as a middleware to send data and commands to the robot, e.g. \cite{kramer_development_2007}. Staranowicz and Mariottini \cite{staranowicz_survey_2011} provided the first comparison of simulation software that used ROS as the communication framework to control a robot. They compared the properties of three open source and commercial simulations. They then demonstrated the capabilities of Gazebo \cite{koenig_design_2004}, a popular simulation software, with a real robot, effectively creating a Digital Twin. However they neither recorded the resources usage nor did they try a different simulator for the real world task. Their work was then extended by Nogueira \cite{nogueira_comparative_2014} who compared 2 simulators and their integration with ROS, the ease of building the world and the CPU usage.

Pitonakova \textit{et al.}~\cite{pitonakova_feature_2018} adopted the methodology in~\cite{nogueira_comparative_2014}. They compared three simulators and then ran extensive tests to record each simulator performance on tasks involving multiple robotic arms. They recorded memory, CPU usage, and the real time factor, meaning the speed at which the simulation runs. This is vital for Digital Twining. It is also vital for machine learning, as the faster the simulation runs without compromising the physics, the faster the training of a machine learning model would be. They performed each tests with and without a Graphical User Interface (GUI) and then compared the impact of rendering the simulation to a screen. Ayala \textit{et al.}~\cite{ayala_comparison_2020} and Korber \textit{et al.}~\cite{korber_comparing_2021} followed the idea of recording resources usage during the running of the experiment. After recapitulating the properties of each simulator, they coded tasks and recorded, memory and CPU usage. Korber \textit{et al.} compared four simulation software on robotic manipulation tasks while Ayala \textit{et al.} only compared 3 for humanoid robot scenarios.

In this paper, we initially consider eight robot simulation software but narrow our benchmark to five that support for ROS2, including two simulation software that have not been considered in the literature. We also propose to implement a pick and place and a throwing tasks to investigate the advantages and limitations for each simulation software, their performance and, ultimately, their suitability for Digital Twins. 

\section{MATERIALS and METHODS} \label{sec:hypothesis}

To evaluate and compare robotic simulation software, we develop our methodology and experiments guided by the following research questions:

\renewcommand{\labelenumi}{\textbf{RQ\arabic{enumi}}}

\begin{enumerate}[leftmargin=1.5cm]
  \item How does simulation software compare in terms of supporting long-term operations while still succeeding at completing a given task?
  \item How repeatable is the simulation software under the same scene and task constrains?
  \item Which simulation software would be more suitable for machine learning research in terms of resource usage and idle time?
\end{enumerate}

\renewcommand{\labelenumi}{\arabic{enumi}}

\subsection{Simulation Software\label{sec:sim_choice}}

The above research questions inform our choice of the simulation software investigated in this paper as shown on Figure \ref{fig:simtable}. Not all of the simulation software have support for ROS2. For this paper, we have attempted to implement our own ROS2 bridge but with limited success due to the rapid development cycle of ROS2. For completeness, we describe our experience while implementing the ROS2 bridge for the simulations we do not use in this paper. Unity's existing bridge is limited as it does not support asynchronous communications which are the underlying communication paradigm in ROS2. Mujuco conflicts with ROS2 because ROS2 multithreaded controller is incompatible with Mujuco single threading nature. Finally, we had to drop Gazebo because development efforts have turned to Ignition, and there is currently an implementation error in the controller code, causing our robot to move erratically\footnote{https://github.com/ros-simulation/gazebo\_ros2\_control/issues/73}.

We also consider simulations that feature a headless mode. This is because, a headless mode is critical in a machine learning context (ref. \textbf{RQ3}). Therefore, we analyse the impact of the GUI in terms of the resource usage. The robot simulation software examined in this paper are: 

\textit{1) Ignition \cite{ignition}} is a set of open source software libraries which are arranged as multiple modular plugins written in Ruby and C++. They have been developed by Open Robotics since 2019. It has a similar communication principle to ROS2. We chose this simulator as it is the successor of Gazebo.

\textit{2)Webots \cite{webots}} has been developed since 1998 by Cyberbotics, a spin-off company from EPFL (Swiss Federal Institute of Technology Lausanne). It supports a wide range of sensors and robot controllers out of the box, as well as being well documented and including several examples files. In Figure \ref{fig:simtable}, it has partial headless support because it only disables the simulation rendering. There is still a GUI visible. We considered it as it is one of the oldest simulation software still being actively developed.

\textit{3) Isaac Sim \cite{isaac}} is a recent, Linux only, simulation environment developed by Nvidia which runs on the PhysX engine, and can be programmed in Python or C. By default, it integrates machine learning capabilities, and has in-built plugins to generate synthetic data for domain adaptation and transfer learning. The latter is possible because of its ray tracing capabilities which allow for a visual simulation as close to reality. While it can be run in headless mode, this is not possible while using their ROS2 extension since there is an issue with the ROS2 plugin not loading when launched from a python script instead of from the terminal.

\textit{4) PyBullet \cite{pybullet}} is a Python-based simulation environment based on the Bullet physics engine which has been in development since 2016. It is popular for machine learning research as it is lightweight and easy to use. For this paper, we implemented a ROS2 compatible plugin since there is no official ROS2 support.

\textit{5) Coppeliasim \cite{vrep}}, previously known as V-REP, is a partially closed source simulation environment, developed since 2010. It can be programmed directly in Lua, or using external controllers in 6 other languages. We decided to include it in our research as it has been compared in previous simulation software reviews, e.g. \cite{nogueira_comparative_2014,pitonakova_feature_2018,melo_analysis_2019,ayala_comparison_2020,ivaldi_tools_2014}. 

\begin{table*}[t]
\centering
\renewcommand{\arraystretch}{1.2}
\caption{Sub-Task overview and Design rationale}
\begin{tabular}{ |p{0.15\linewidth}|p{0.4\linewidth}|p{0.08\linewidth}|p{0.25\linewidth}| }
\hline
Name & Design & Features Tested & Data Recorded and Rationale\\
\hline
Task 1-A: Pick and Place & A robotic arm randomly takes 5 cm cubes from a table with 21 cubes arranged in a $7\times3$ grid. The task aim is to stack them into 3 towers of 5 cubes as can be seen in Figure \ref{fig:overview}. We consider 3 stacks in order to leave cubes on the table and to allow for more diversity in each repetition. We set the limit to 5 stacked cubes due to the height of the table and the capabilities of the robot. & \multirow{2}{\linewidth}[-3pt]{Friction, Gravity, Inertia}& \multirow{2}{\linewidth}[-3pt]{This experiment addresses \textbf{RQ2} which analyses the numbers of cubes correctly placed. It will also test the stability and suitability of the simulation for long operations, as recorded by the number of cubes still in place at the end (ref. \textbf{RQ1}).
The idea of stacking cubes to analyse performance is motivated from ~\cite{korber_comparing_2021}}. \\
\cline{1-2}
Task 1-B: Pick and Place Headless & We use the same setup as Task 1-A but without a GUI. This was chosen as in a machine learning setting, experiments need to be restarted multiple times and often run on a server with no screen (ref. \textbf{RQ3}). & & \\
\hline
Task 2-A: Throwing & A robotic arm will pick up a cube and throw it towards a pyramid of 6 cubes. The arm goes as far back as mechanically possible and perform a throwing motion towards the pyramid in front of it. Figure \ref{fig:overview} shows the trajectory taken by the robot during this task. The cube is released at $50\%$ of the trajectory. The pyramid is placed such that a successful throw at full power will collapse it. & \multirow{2}{\linewidth}[-3pt]{Friction, Gravity, Inertia, Kinetic} & \multirow{2}{\linewidth}[-3pt]{This task benchmarks the accuracy and repeatability of the simulation software and addresses \textbf{RQ2}. The latter is carried out by recording the number of cubes that are displaced from their original position. This idea has been inspired by a contributor to Ignition\footnotemark{} demonstrating how to interface ROS2 and Ignition.} \\
\cline{1-2}
Task 2-B: Throwing Headless & We follow the same design as Task 2-A, except without a GUI (ref. \textbf{RQ3}). & &\\
\hline
\end{tabular}
\label{table:Task_overview}
\end{table*}

\subsection{Data Capturing}

For data capturing and recording, we adopt the metrics mentioned in Section \ref{sec:background}, namely, the processor usage (CPU), the memory usage (RAM) and the execution time (ref. \textbf{RQ1 \& 3}). We also record task-specific data, such as the number of cubes placed or moved (ref. \textbf{RQ2 \& 3}). The execution time is not mentioned in the literature but was added as a metric for machine learning, in which the running time can have major impact, as training a model involves several iterations of the simulation. A delay of tens of seconds for one iteration can turn into hours of difference for long training sessions. In order to accurately record each simulation, we start recording 5 seconds before the simulation starts and ends the recording 60 second after the task has ended. We record processes only related to the simulation while discarding the rest, such OS-specific processes. 

\subsection{Robotic Tasks} \label{sec:Task}

We consider 2 tasks, each divided into 2 sub-tasks, in order to evaluate each simulator considered here. A sub-task is repeated 20 times with the aim to reduce the variance during data recording and to obtain an accurate statistical characterisation of a simulation. In practice, we found that more than 20 repetitions does not result in a statistical significant difference. The 2 tasks along with their rationale are summarised in Table \ref{table:Task_overview}. The task execution logic is the same for all simulations. We must note that we use the default simulation parameters to setup these tasks. This is to remove bias while implementing the tasks and avoid tuning simulator specific parameters in order to obtain an objective evaluation for each simulation software.

\subsection{Robot Control}

There are 3 methods to control a robot using ROS2, namely, the joint controller, the joint trajectory follower and the general purpose ROS controller. The joint controller sets the position of the joints to a given joint angle using hardware specific interfaces of a robot. This is the simplest method as it provides no feedback to the controller. The joint trajectory follower uses a ROS action client/server combination in which the client sends the joint position for a given trajectory as a list along with a delay. Then, the server continuously sends the current value of the joints as a feedback mechanism until the trajectory is completed. This method works well in practice and we have implemented it for Coppeliasim, PyBullet and Isaac Sim. For the Ignition and Webots, we use the general purpose ROS controller (\texttt{ros\_control})  \cite{chitta_ros_control_2017}, which is not implemented for the other simulations. It provides a wrapper for the joint trajectory follower described above, but also provides different methods of control such as a velocity, effort or gripper controller.

\section{EXPERIMENTS\label{sec:experiments}}

\subsection{Methodology}

We use a docker container with Nvidia Ubuntu 20.04 \texttt{cudagl} image for all simulators except for Isaac Sim that cannot access the base graphics driver API when using docker. Isaac Sim is thus executed in the base system from where we run all experiments. ROS2 Foxy has been installed, along with simulator specific packages. Docker has been used to easily provide an image with all the necessary packages installed without conflict between different simulations. It also allows for reproducibility of these experiments by providing the same setup every time. The base system runs an Intel I7 10700 with 32GB of RAM and an Nvidia GeForce RTX 3060 with Ubuntu 20.04. 
\footnotetext[2]{https://github.com/AndrejOrsula/ign\_moveit2}
We used \texttt{psutil}\footnote{https://pypi.org/project/psutil/} which is a python package that records the CPU and RAM usage. Each process was monitored at 10 Hz to minimise the resources impact. For each simulator we used the recommended time step, and we have fixed all simulators to run in real time. We use the Franka Panda robot, and its model and configuration files provided by MoveIt 2~\cite{coleman_david_reducing_2014}. The project repository can be found at \url{https://github.com/09ubberboy90/ros2_sim_comp}.

\begin{figure*}[t]
\centering
\includegraphics[width=0.85\textwidth]{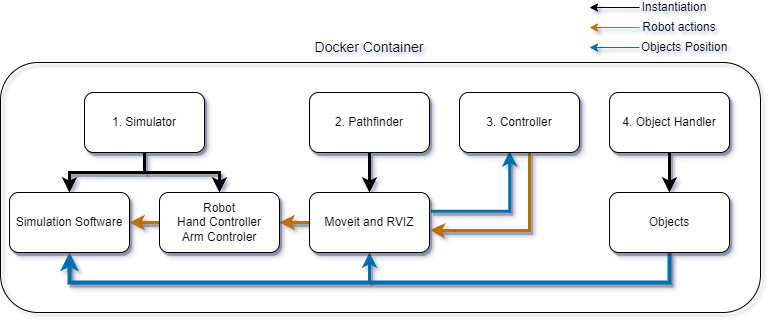}
\caption{Implementation Diagram}
\label{fig:Implementation}
\end{figure*}

\subsection{Implementation} \label{sec:implementation}

The implementation comprises 4 components as shown in Figure \ref{fig:Implementation} and as noted below.

\begin{enumerate}
    \item \textit{Simulator:} launches the simulation software and spawns the robot along with the gripper and arm controllers.
    \item \textit{Pathfinder:} launches Rviz (a ROS data visualisation software) and MoveIt 2.
    \item \textit{Controller:} chooses which object to grab from the list of collision objects advertised by MoveIt 2, shown in Figure \ref{fig:Implementation} as blue arrows. Then, the pick and place or throwing task is executed accordingly.
    \item \textit{The Object-Handler:} spawns the objects and publishes their position to the planning scene of MoveIt 2 at 2 Hz. We choose 2 Hz because the scene's rate of change in real time does not deviate considerably. Higher rates consume more CPU usage which impacts the performance of all simulations in this paper.
\end{enumerate}

In our implementation, both the arm and the gripper are controlled using a position controller. We must note that the gripper controller available in ROS1 has not yet been ported to ROS2. The latter causes issues while grasping objects in the simulations (except Webots) as the amount of force is constant with no feedback. To mitigate this issue, we command the gripper to close its gripper beyond the optimal closing distance to ensure that the object is grasped correctly. Webots does not have this issue because it implements PID controller by default for each simulated motor.
These 4 components are launched using a single ROS2 launch file with ad hoc delays to ensure everything is started as Part 3 and 4 do not work unless part 2 is launched. For each simulation software, we are using the default physics time step. The physics engine is also the default, except for Coppeliasim, in which we use the \textit{Newton physics engine} because the other supported physics engines causes the gripper to fail to grip the cube.

\subsection{Task 1 Experiments}\label{sec:t1}

\begin{table}[t]
\renewcommand{\arraystretch}{1.1}
\centering
\caption{Task 1 results}
\label{table:T1_Results}
\setlength\tabcolsep{4pt} 
\begin{tabular}{ |l|c|c|c|c|c|c|c| }
\hline
\multirow{3}{*}{Name} & \multirow{3}{0.1\linewidth}{\centering Failure (\%)} & \multicolumn{3}{c|}{Cubes Placed} & \multicolumn{3}{c|}{Cubes at the end}\\
 & & \multicolumn{3}{c|}{ (\%)} & \multicolumn{3}{c|}{of the task (\%)}\\
\cline{3-8}

&  & Min&Mean&Max& Min&Mean&Max\\
\hline
Ignition & 0 & 60 & 91 & 100 & 47 & 82 & 100 \\
Ignition GUI & 0 & 80 & 94 & 100 & 67 & 88 & 100 \\
Isaac Sim GUI & 10 & 47 & 89 & 100 & 20 & 65 & 100 \\
Pybullet & 0 & 7 & 18 & 47 & 7 & 15 & 47 \\
Pybullet GUI & 0 & 7 & 11 & 40 & 7 & 11 & 40 \\
Coppeliasim & 30 & 67 & 92 & 100 & 47 & 79 & 100 \\
Coppeliasim GUI & 15 & 53 & 91 & 100 & 47 & 76 & 100 \\
Webots & 5 & 53 & 88 & 100 & 47 & 79 & 93 \\
Webots GUI & 5 & 73 & 91 & 100 & 40 & 80 & 100 \\

\hline
\end{tabular}
\end{table}

Table \ref{table:T1_Results} shows the result of task 1, which addresses \textbf{RQ1 \& RQ2}. The reason the task times out (failure in Table \ref{table:T1_Results}) is because the ROS controller fails to start, or in the case of Coppeliasim, the client refuses to connect to the simulation server. The rest of the metrics only focus on the successful attempts.

Ignition and PyBullet did not have timeouts; however, PyBullet performs significantly worse at stacking 5 towers than the other simulators as $15\%$ of the cubes in average (i.e 3 cubes) were correctly positioned at the end of the simulation, and, therefore, the robot does not encounter scenarios where it collapses towers. Ignition and Webots are the best performing simulations at executing the task of stacking cubes, and at keeping the cubes in place. Coppeliasim and Isaac Sim, are carrying out the task well at placing the cube in the right place but, tend to have situations where the robot collapses the towers. Furthermore, while Coppeliasim achieves $92\%$ success of placing cubes, we can observe that it often times out, and reduces its overall success. We can also observe in Table \ref{table:T1_Results} that there is no a statistical significant difference between headless and GUI modes. These results suggest that Ignition (headless and GUI) succeeds at completing the task more frequently using the default parameters (ref. \textbf{RQ1}) and has less variation over different attempts (ref. \textbf{RQ2}).

\begin{table}[t]
\centering
\renewcommand{\arraystretch}{1.15}
\caption{Task 1 Resources usage.}
\label{table:T1_Usage}
\begin{tabular}{ |l|c|c|}
\hline
Name & CPU (\%) & RAM (MB) \\
\hline
Ignition & 202 ± 94 & 686 ± 313 \\
Ignition GUI & 205 ± 51 & 775 ± 159 \\
Isaac Sim GUI & 134 ± 25 & 10070 ± 1885 \\
Pybullet & 117 ± 41 & \textbf{663 ± 224} \\
Pybullet GUI & 140 ± 28 & 919 ± 168 \\
Coppeliasim & 135 ± 42 & 860 ± 262 \\
Coppeliasim GUI & \textbf{100 ± 43} & 850 ± 379 \\
Webots & 144 ± 72 & 1191 ± 550 \\
Webots GUI & 162 ± 61 & 1322 ± 410 \\
\hline
\end{tabular}
\end{table}

Table \ref{table:T1_Usage} shows that PyBullet headless consumes fewer resources overall, while Isaac Sim, is the most memory intensive simulation as it consumes 10 times more RAM than the next simulator (Webots GUI). This is inline with the current trend of Pybullet being used in machine learning research (ref. \textbf{RQ3}). It is worth noting that Coppeliasim uses fewer resources with a GUI than headless. We speculate that this is because it was initially designed as a GUI application, with headless support only added at a later date, thus having received less development focus.

\begin{figure}[t]
\includegraphics[width=0.45\textwidth]{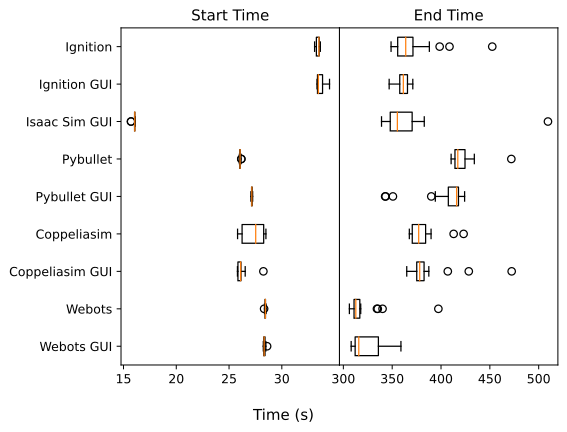}

\caption{Task 1: Mean task start and end time}
\label{fig:T1_sim_start}
\end{figure}

Figure \ref{fig:T1_sim_start} shows the spread of the start time and end time for each simulation (ref. \textbf{RQ3}). As mentioned in \ref{sec:implementation}, Isaac Sim has to be started manually, thus the time that takes the simulation to start is not captured in the plot. Ignition takes the most time to load, because it requires an additional movement to the start position that it has to do to start the effort controller. Webots finishes the earliest with little variation. Combined with its relatively high success rate from Table \ref{table:T1_Results}, Webots appears to be ideal for prototyping robotic tasks and for machine learning experiments due to its relatively high success rate from Table \ref{table:T1_Results} and finishing the task and simulation early with low variation. 
PyBullet, on the other hand, takes the most time, and combined with its high failure rate (with the default parameters), it may not be suitable for machine learning as it would take more time to run a single experiment. Similarly, further parameter tuning would be required in order to obtain a stable simulation that succeeds at completing the task.

\subsection{Experiment 2}\label{sec:t2}

\begin{table}[t]
\renewcommand{\arraystretch}{1.1}
\centering
\caption{Task 2 results}
\label{table:T2_Results}
\setlength\tabcolsep{4pt} 
\begin{tabular}{ |l|c|c|c|c| }
\hline
\multirow{2}{*}{Name} & \multirow{2}{0.1\linewidth}{\centering Failure (\%)} & \multicolumn{3}{c|}{Cubes Moved \%)}\\
\cline{3-5}

&  & Min&Mean&Max\\
\hline
Ignition & 10 & 0 & 0 & 0 \\
Ignition GUI & 0 & 0 & 0 & 0 \\
Isaac Sim GUI & 0 & 0 & 18 & 83 \\
Pybullet & 0 & 0 & 0 & 0 \\
Pybullet GUI & 0 & 0 & 0 & 0 \\
Coppeliasim & 32 & 0 & 9 & 50 \\
Coppeliasim GUI & 20 & 0 & 0 & 0 \\
Webots & 5 & 0 & 11 & 50 \\
Webots GUI & 15 & 0 & 21 & 50 \\
\hline
\end{tabular}
\end{table}

As shown in Table \ref{table:T2_Results}, which focuses on \textbf{RQ1 \& 3}, only Webots throws consistently. Isaac Sim consistently manages to throw the cube but fails to hit the pyramid as the motion behaviour is not consistent. We speculate that this is because we did not tune the simulation parameters and used the default values. Coppeliasim and PyBullet manages to hit the pyramid, but the behaviour is rare as the few times the arm manages to successfully perform the throwing motion, the throw is not always at the same distance nor perfectly aligned. Coppeliasim has a high timing out rate (failure column in Table \ref{table:T2_Results}) due to the reasons mentioned in Section \ref{sec:t1}. Finally, for Ignition, the success at hitting the pyramid is zero. We observe that, in most cases, the cube falls in transit, especially when the arm is as far back as possible and starts to move at full speed for the throwing motion. At this point, the robot and the cube have the highest moment of inertia, and if the friction between the cube and the gripper is not enough, the cube falls. We must note that we fix the friction parameter to explore the default capabilities for each simulator. We also notice that there are instances when the robot manages to throw the cube but does not hit the pyramid. This is because the gripper controller had a delay in opening its gripper, changing the thrown cube landing spot.

\begin{table}[t]
\renewcommand{\arraystretch}{1.15}

\centering
\caption{Task 2 Resources usage}
\label{table:T2_Usage}
\begin{tabular}{ |l|c|c|}
\hline
Name & CPU (\%) & RAM (MB) \\
\hline
Ignition & 152 ± 99 & \textbf{524 ± 340} \\
Ignition GUI & 137 ± 103 & 574 ± 371 \\
Isaac Sim GUI & 148 ± 48 & 11123 ± 1865 \\
Pybullet & 139 ± 32 & 730 ± 174 \\
Pybullet GUI & 136 ± 31 & 828 ± 205 \\
Coppeliasim & 115 ± 37 & 742 ± 185 \\
Coppeliasim GUI & \textbf{98 ± 28} & 838 ± 214 \\
Webots & 141 ± 74 & 1275 ± 261 \\
Webots GUI & 134 ± 72 & 1207 ± 198 \\
\hline
\end{tabular}
\end{table}

\begin{figure}[t]
\includegraphics[width=0.45\textwidth]{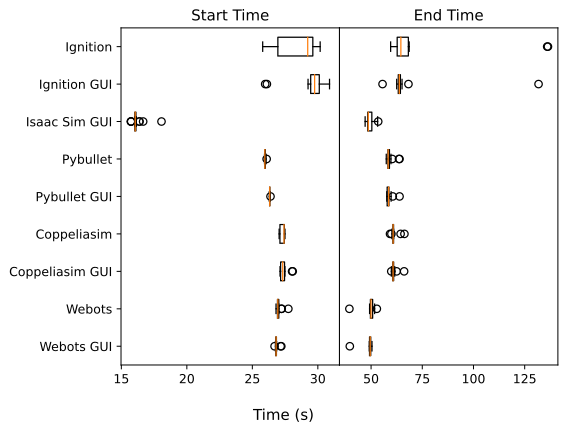}

\caption{Task 2: Mean task start and end time}
\label{fig:T2_sim_start}
\end{figure}

Table \ref{table:T2_Usage} shows similar results to task 1. Coppeliasim uses the lowest amount of CPU while Ignition uses the less memory. The CPU usage for all simulations observes less variation. This could be due to the simplicity of the world and the short time of execution. As mentioned in \ref{sec:t1}, Coppeliasim still uses fewer resources with a GUI than headless. Figure \ref{fig:T2_sim_start} shows similar start and end time for all simulations, observing lower variations compared to task 1. The reason for this is because the relatively short time of execution and the low amount of path planning that can fail and delay the execution. For this scenario, considering only the time of execution will not have and impact on the choice for a machine learning approach as the difference between execution is minimal. If the resource usage is important, then Coppeliasim should be considered for machine learning tasks. Otherwise, a more successful simulation should be considered such as Webots.

\section{CONCLUSIONS \& FUTURE WORK}


In this paper, we have investigated current robot simulation software performance and their suitability to perform two different robotic manipulation tasks. We have also developed a methodology to systematically benchmark robot simulations under similar parameters, tasks and scenarios. Based on our experimental results, Webots appears to be the more suitable for long-term operations while still succeeding at completing a given task (ref. \textbf{RQ1}) and be able to replicate the same simulation conditions across attempts (ref. \textbf{RQ2}). Webots would only be suitable for machine learning if the execution time and resources are not a requirement while training machine learning models (ref. \textbf{RQ3}). Ignition, while comparable to Webots, is more suited to answer \textbf{RQ1 \& RQ3}. \textbf{RQ2} is only satisfied if the task is slow moving and constant. We must note that Ignition is still in development and some of the challenges we encountered while implementing both tasks and carrying out our experiments may be mitigated in the future. Coppeliasim and PyBullet have less impact in terms of resource usage and are the most suited to answer \textbf{RQ3}. That is, Coppeliasim provides better stability for task success at the cost of timing out more often. Finally, Isaac Sim only satisfies \textbf{RQ1}, as the simulated scene was not repeatable across attempts. 

From our review and experimental results, we found that current robot simulation software could not be used to develop a digital twin. This is because the simulators considered in this paper cannot maintain a repeatable simulated scene over time. We hypothesise that a continuous feedback mechanism is needed between the simulation and reality similar to \cite{chang_sim2real2sim_2020} in order to maintain an accurate representation of the real environment. While this paper focused on benchmarking robot simulation software, future work consists of optimising each simulator to minimise failure rates and maximise task completion, and benchmark them accordingly. 
Additionally, the Unreal Engine plugin for ROS2, has recently seen more development and could potentially replace Unity in our original plan. We also aim to specifically benchmark each simulation in a machine learning context such as in \cite{pitsillos_intrinsic_2021} with the view to develop a digital twin that can take advantage of a simulated environment to deploy AI solutions for autonomous robotic systems.


\bibliographystyle{IEEEtran}
\bibliography{zotero, References}

\end{document}